\documentclass{article}

     \PassOptionsToPackage{numbers, compress}{natbib}
\usepackage{arxiv}
\usepackage[utf8]{inputenc} % allow utf-8 input
\usepackage[T1]{fontenc}    % use 8-bit T1 fonts
\usepackage{hyperref}       % hyperlinks
\usepackage{url}            % simple URL typesetting
\usepackage{booktabs}       % professional-quality tables
\usepackage{amsfonts}       % blackboard math symbols
\usepackage{nicefrac}       % compact symbols for 1/2, etc.
\usepackage{microtype}      % microtypography
\usepackage{url}
\usepackage{graphicx}
\usepackage{algorithm}
\usepackage{algcompatible}
\usepackage{amsfonts}
\usepackage{amssymb}
\usepackage{amsthm}
\usepackage{amsmath}
\usepackage{bbm}
\usepackage{float}
\usepackage[font={rm,md,up}]{subfig}
\usepackage{caption,stackengine}
\usepackage{ragged2e}
\usepackage{tikz}
\usetikzlibrary{positioning}
\usetikzlibrary{arrows.meta}
\usepackage{multirow}
\usepackage{arydshln}

\makeatletter
\newcommand\footnoteref[1]{\protected@xdef\@thefnmark{\ref{#1}}\@footnotemark}
\makeatother

\tikzset{
  >={Latex[width=1mm,length=2mm]}
}

\tikzset{every picture/.style={line width=0.75pt}} %set default line width to 0.75pt    

\newcommand\blfootnote[1]{%
  \begingroup
  \renewcommand\thefootnote{}\footnote{#1}%
  \addtocounter{footnote}{-1}%
  \endgroup
}

\DeclareMathOperator*{\argmin}{argmin}

\title{Generalizing to unseen domains via distribution matching}

\author{Isabela Albuquerque\textsuperscript{1,}\thanks{Correspondence to \texttt{isabelamcalbuquerque@gmail.com}} , Jo\~ao Monteiro\textsuperscript{1}, Mohammad Darvishi\textsuperscript{2}, \textbf{Tiago H. Falk\textsuperscript{1}, Ioannis Mitliagkas\textsuperscript{3}}   
\\
\textsuperscript{1}INRS-EMT, Universit\'e du Qu\'ebec\\
\textsuperscript{2} Faubert Lab, Universit\'e de Montr\'eal \\ 
\textsuperscript{3}Mila \& DIRO, Universit\'e de Montr\'eal} 

\begin{document}

\maketitle

\begin{abstract}

Supervised learning results typically rely on assumptions of i.i.d. data. Unfortunately, those assumptions are commonly violated in practice. In this work, we tackle such problem by focusing on domain generalization: a formalization where the data generating process at test time may yield samples from never-before-seen domains (distributions).
Our work relies on the following lemma: by minimizing a notion of discrepancy between all pairs from a set of given domains, we also minimize the discrepancy between any pairs of mixtures of domains. Using this result, we derive a generalization bound for our setting. We then show that low risk over unseen domains can be achieved by representing the data in a space where (i) the training distributions are indistinguishable, and (ii) relevant information for the task at hand is preserved. Minimizing the terms in our bound yields an adversarial formulation which estimates and minimizes pairwise discrepancies.
We validate our proposed strategy on standard domain generalization benchmarks, outperforming a number of recently introduced methods.
Notably, we tackle a real-world application where the underlying data corresponds to multi-channel electroencephalography time series from different subjects, each considered as a distinct domain.

 %The proposed strategy achieves strong performance in standard domain generalization benchmarks outperforming a number of recently introduced methods. We further tackle a more practical setting where the underlying data corresponds to physiological signals from patients where each such patient is viewed as a domain.

\end{abstract}

\section{Introduction}
The main assumption within the empirical risk minimization framework is that all examples used for training and testing predictors are independently drawn from a fixed distribution, i.e. the i.i.d. assumption. A number of generalization guarantees were derived upon that assumption and those results induced several algorithms for the solution of supervised learning problems. However, important limitations in this setting can be highlighted: i) the i.i.d. property is \emph{unverifiable} \cite{langford2005tutorial} given that one doesn't have access to the data distribution, and  ii) it doesn't account for distribution shifts which often occur in practice. Representative examples of these distribution shifts include changes in data acquisition conditions, such as illumination in images for object segmentation, or new data sources such as unseen speakers when performing speech recognition.

A number of alternative settings was then introduced in order to better cope with more realistic cases. Risk minimization under the \emph{domain adaptation} setting, for instance, relaxes part of the i.i.d. assumption by allowing a source distribution (or domain)\footnote{We use the terms \textit{domain}, \textit{data distribution}, and \textit{data source}  interchangeably throughout the text.} \blfootnote{\textbf{This work has been submitted to the IEEE for possible publication. Copyright may be transferred without notice, after which this version may no longer be accessible.}} as well as a different target distribution observed at test time. The domain adaptation results introduced in \cite{ben2007analysis} showed that the generalization gap in terms of risk difference across the two considered distributions for a fixed predictor is upper bounded by a notion of distance measured between the training and testing domains. While less restrictive than the previous setting, the domain adaptation case is still limited in that only pairs of distributions seen during training are expected to yield low risk, and shifts beyond those domains will likely induce poor performance. Moreover, algorithms devised for this setting \emph{rely on access at training time to an unlabeled sample from the target distribution} so that representations can be learned inducing invariance across train and target domains \cite{ganin2016domain}. This is a limiting factor for practical applications where target domain data may be inaccessible; for example, a speech recognition service cannot be (re)trained on data obtained from every new speaker it observes.

A more general setting is often referred to as \textit{domain generalization} \cite{muandet2013domain}. In this case, it is assumed that a set of distributions over the data is available at training time. At test time, however, both observed distributions as well as unseen novel domains might appear, and a low risk is expected regardless of the underlying domain. More importantly, unlike domain adaptation in which the goal is to find a representation that aligns training data distributions with a specific target domain, \emph{domain generalization strategies aim at finding a representation space that yields good performance on novel distributions, unknown at training time.} Recent work on domain generalization has included the use of data augmentation \cite{shankar2018generalizing, volpi2018generalizing} at training time, meta-learning to simulate domain shift \cite{li2018learning}, adding a self-supervised task to encourage an encoder to learn robust representations \cite{carlucci2019domain, albuquerque2020improving}, and  learning domain-invariant representations \cite{li2018deep}, among other approaches.

In this paper, we propose an innovation within the domain generalization setting. We first argue and prove that, given a set of distributions over data, if the distances measured between any pair of such distributions is small, so is the distance between mixtures obtained from the same set. This leads to the development of a bound on the risk measured against any distribution, and further shows that generalization can be expected if one considers distributions on the neighborhood of the ``convex hull''\footnote{i.e., the set of all mixtures obtained from given distributions.} defined by the set of domains accessible during training. Inspired by these findings, we define an approach so that an encoder is enforced to map the data to a space where domain-dependent cues are filtered away while relevant information to the task of interest is conserved. While doing so, unlike standard domain adaptation approaches, no data from test distributions is observed. 

We summarize our contributions in the following:
\begin{enumerate}
    \item We introduce assumptions on the data generating process tailored to the domain generalization setting, which we argue are more general than standard i.i.d. requirements and more likely to hold in practice. In other words, given a data sample, it is more likely that our assumptions will hold compared to the more restrictive i.i.d. property;
    \item We prove a generalization bound for the risk over unseen domains and show that generalization can be expected for domains on the neighborhood of a notion of convex hull of distributions observed at training time;
    \item Aiming to minimize the terms of the introduced bound, we devise an adversarial approach so that pairwise domain divergences are estimated and minimized. In order to do so, several practical improvements are proposed on top of previous approaches for domain adaption including the use of random projection layers prior to domain discriminators.
    \item We provide evidence through empirical evaluation showing that the proposed approach yields improvements relative to alternative methods across scenarios where different assumptions over the observed domains hold, including realistic cases where the labeling functions might shift.
\end{enumerate}

The remainder of this paper is organized as follows: In Section \ref{sec:background} we introduce the notation adopted within this work, and review related work and generalization guarantees. In section \ref{sec:method}, we define the domain generalization setting and present our main results, as well as the resulting algorithm. Section \ref{sec:res} provides the experiment descriptions and their respective results. Section~\ref{sec:conc} concludes the paper.

\section{Background}\label{sec:background}
\subsection{Notation}
Let the data be represented by $\mathcal{X} \subset \mathbb{R}^D$, while $\mathcal{Y}$ denotes the label space. Examples correspond to a pair $(x, y) : x \in \mathcal{X}, y \in \mathcal{Y}$, such that $y=f_{\mathcal{D}}(x)$, and $f_{\mathcal{D}}:\mathcal{X}\rightarrow\mathcal{Y}$ is a deterministic labeling function. 

A domain is defined as a tuple $\langle \mathcal{D}, f_{\mathcal{D}} \rangle$ where $\mathcal{D}$ corresponds to a probability distribution over $\mathcal{X}$. Moreover, we define a mapping $h:\mathcal{X}\rightarrow\mathcal{Y}$, such that $h \in \mathcal{H}$, where $\mathcal{H}$ is a set of candidate hypothesis, and finally define the risk $R$ associated with a given hypothesis $h$ on domain $\langle \mathcal{D}, f_{\mathcal{D}} \rangle$ as:
\begin{equation}
    R[h] = \mathbb{E}_{x \sim \mathcal{D}} \ell [h(x), f_\mathcal{D}(x)],
\end{equation}
where the loss $\ell:\mathcal{Y} \times \mathcal{Y} \rightarrow R_{+}$ quantifies how different $h(x)$ is from the true labeling function $y=f_{\mathcal{D}}(x)$ for a given instance $(x, y)$.

\subsection{Related work}\label{sec:rel_work}
A number of contributions under the domain generalization setting borrowed tools from causal inference to enforce the learned representations to be invariant across the different domains presented to the model at training time \cite{arjovsky2019invariant,mahajan2020domain,ahuja2020invariant}. Other contributions such as \cite{carlucci2019domain} and \cite{albuquerque2020improving} proposed different strategies to leverage self-supervised tasks to improve the out-of-distribution performance of a given model.
Inspired by the domain adaptation literature \cite{kifer2004detecting,ben2007analysis,ganin2016domain}, previous work on domain generalization also proposed to add a regularization term based on the minimization of a notion of divergence between the source domains to the empirical loss computed on the training data. This is the case of CIDDG \cite{li2018deep}, where class-specific domain classifiers are employed to induce the encoder to learn representations where the mismatch between the labels conditional distributions is minimized. Moreover, \cite{li2018domain} proposed MMD-AAE, an approach that relies on an adversarial autoencoder used along with a maximum mean discrepancy penalty \cite{gretton2012kernel} to remove domain-specific information. 

Recent work has proposed settings where domain-shifts are simulated at training time by splitting the source domains into meta-train and meta-test sets \cite{li2018learning,balaji2018metareg,dou2019domain,li2019episodic}. Strategies based on learning domain-invariant representations \cite{muandet2013domain}, data augmentation \cite{volpi2018generalizing,shankar2018generalizing}, and on decomposing the model's parameters into domain-agnostic and domain-specific components \cite{li2017deeper} have also been introduced. Work on other settings with more restrictive assumptions than domain generalization are also related to our contribution. For example, recent work on multi-domain learning \cite{schoenauer2019multi}, a setting where multiple domains are available at training time and test data is drawn from the same distributions seen during training \cite{dredze2010multi}, also leveraged an adversarial approach to perform $\mathcal{H}$-divergence minimization.

% In the domain generalization setting, one's interest is to train a model $h$ able to minimize the risk on unseen data distributions. Given $N_S$ data sources, we assume that examples $\{x^j_m, y^j_m\}_{m=1}^{M_j}$ for each domain $j\in\{1, \ldots, N_S\}$ consist of the only data available during training.

A straightforward approach to extend the empirical risk minimization (ERM) setting for domain generalization would be to learn $h$ minimizing the empirical risk $\hat{R}[h]$ measured over all $N_S$ source domains and \textit{hope} generalization would be achieved to the target data, i.e.:
\begin{equation}\label{ERM}
     h = \arg \min \hat{R} = \frac{1}{N_S} \sum_{j=1}^{N_S}\frac{1}{M_j}\sum\limits_{i=1}^{M_j} \ell[h(x_i),f(x_i)].
\end{equation}
In fact, as will be discussed in more detail in next sections, such a rather simplistic approach often yields strong baselines.

% In the domain generalization setting, one's interest is to train a model $h$ able to minimize the risk on unseen data distributions. Given $N_S$ data sources, we assume that examples $\{x^j_m, y^j_m\}_{m=1}^{M_j}$ for each domain $j\in\{1, \ldots, N_S\}$ consist of the only data available during training.

% One trivial approach to extend the empirical risk minimization (ERM) setting for domain generalization would be to learn $h$ minimizing the empirical risk $\hat{R}[h]$ measured over all $N_S$ source domains and \textit{hope} generalization would be achieved to the target data, i.e.:

% \begin{equation}\label{ERM}
%     h = \arg \min \hat{R} = \frac{1}{N_S} \sum_{j=1}^{N_S}\frac{1}{M_j}\sum\limits_{i=1}^{M_j} \ell[h(x_i),f(x_i)].
% \end{equation}

% In fact, as will be discussed in more detail in next sections, such a rather simplistic approach often yields strong baselines.

\subsection{Generalization guarantees for domain adaptation} 
We now state results from the domain adaptation literature which are relevant in the context of this work. 
The discussion in \cite{ben2010theory} established the theoretical foundations for studying cross-domain generalization properties for domain adaptation problems. They showed that, given a source domain $\mathcal{D}_S$ and a target domain $\mathcal{D}_T$, the risk of a given hypothesis $h \in \mathcal{H}$, $h:\mathcal{X}\rightarrow\{0,1\}$, on the target is bounded by:
\begin{equation}\label{eq:bound_da}
    R_T[h] \leq R_S[h] + d_{\mathcal{H}\Delta\mathcal{H}}[\mathcal{D}_S, \mathcal{D}_T] + \lambda,
\end{equation}
where $\lambda$ corresponds to the minimal total risk over both domains which can be achieved within a given hypothesis class $\mathcal{H}$. The term $d_{\mathcal{H}\Delta\mathcal{H}}[\mathcal{D}_S, \mathcal{D}_T]$ corresponds to the $\mathcal{H}$-divergence introduced in \cite{kifer2004detecting} for a hypothesis class $\mathcal{H}\Delta\mathcal{H}=\{h(x) \oplus h'(x) | h, h' \in \mathcal{H}\}$, where $\oplus$ is the XOR function. The $\mathcal{H}$-divergence between two distributions $\mathcal{D}_S$  and $\mathcal{D}_T$ is defined as: 
\begin{equation}
        d_{\mathcal{H}}[\mathcal{D}_S, \mathcal{D}_T] = 2 \sup_{\eta \in \mathcal{H}} | \text{Pr}_{x\sim\mathcal{D}_S} [\eta(x)=1]
        - \text{Pr}_{x\sim\mathcal{D}_T} [\eta(x)=1] |.        
\end{equation}
%where $\eta: \mathcal{X} \rightarrow \{0, 1\}$ is a discriminator responsible for distinguishing examples from $\mathcal{D}_S$ and $\mathcal{D}_T$.
As discussed in \cite{ben2007analysis}, an estimate of $d_{\mathcal{H}}[\mathcal{D}_S, \mathcal{D}_T]$ can be directly computed from the error of a binary classifier trained to distinguish domains.

\section{Learning domain agnostic representations for domain generalization}\label{sec:method}

\subsection{Formalizing domain generalization}

We start by defining a set of assumptions over the data generating process considering the domain generalization case as well as the notion of risk we are concerned with. We then define $\mathfrak{D}$, referred to as \emph{meta-distribution}, corresponding to a probability distribution over a countable set of possible domains. Under this view, a query for a data example consists of: i) sampling a domain from the meta-distribution, and ii) sampling a data point according to that particular domain. Such process is repeated $m$ times so as to yield a training sample $(x^m\sim \mathfrak{D}^m, y^m)$. We remark the described model of data generating processes is sufficiently general so as to include the i.i.d. case (the meta-distribution yields a single domain) as well as the domain adaptation setting (if two domains are allowed), but further supports several other cases where multiple domains exist.

Once a finite train sample is collected, a set of $N_S$ domains is observed. Each distribution $\mathcal{D}^i_S$, $i \in [N_S]$, in such set will be referred to as source domain. At test time, however, drawing samples from $\mathfrak{D}$ might yield data distributed according to new unseen domains. We then introduce extra notation and represent the set of possible domains unobserved while train data is acquired by $\mathcal{D}^j_U$, $j \in [N_U]$. The labeling rules corresponding to each domain are denoted as $f_{S_i}$ and $f_{U_j}$, for the source and unseen domains, respectively. For the sake of clarity, we hereinafter omit the index from the notation corresponding to unseen domains whenever it can be inferred from the context.
%\end{minipage}
%\end{minipage}

We proceed and define a risk minimization framework similar to that corresponding to the i.i.d. setting: find the predictor $h^* \in \mathcal{H}$ that minimizes the meta-risk $R_{\mathfrak{D}}[h]$ defined as follows: 
\begin{equation}\label{eq:risk_DG}
\begin{split}
& h^* \in \argmin_{h\in \mathcal{H}} \, R_{\mathfrak{D}}[h], \\  
& R_{\mathfrak{D}}[h]=\mathbb{E}_{\mathcal{D}\sim \mathfrak{D}}[\mathbb{E}_{x\sim\mathcal{D}}[\ell(h(x), f_{\mathcal{D}}(x))]].    
\end{split}
\end{equation}
However, within the domain generalization setting, no information regarding possible test distributions is available at training time, which renders estimating $R_{\mathfrak{D}}[h]$ uninformative for a practical number of source domains. Moreover, we argue that no-free-lunch type of impossibility results may be used to conclude that it is impossible to generalize to any possible unknown distribution\footnote{For a fixed $h$, one can always define a distribution yielding high risk.}, so that one must assume something about the test domains in order to enable generalization. In the following results, we tackle this issue and introduce generalization guarantees for a particular set of domains lying close to the set of mixtures of \emph{source distributions}, i.e., those observed once train data is collected.

\subsection{Matching distributions in the convex hull}
Let a set $S$ of source domains such that $|S|=N_S$ be denoted by $\mathcal{D}^i_S$, $i \in [N_S]$. The convex hull $\Lambda_S$ of $S$ is defined as the set of mixture distributions given by: $\Lambda_S = \{\bar{\mathcal{D}}: \bar{\mathcal{D}}(\cdot) = \sum_{i=1}^{N_S} \pi_{i} \mathcal{D}^i_S(\cdot), \pi_{i} \in \Delta_{N_S-1}\}$, where $\Delta_{N_S-1}$ is the $N_S-1$-th dimensional simplex. The following lemma shows that for any pair of domains such that $\mathcal{D}', \mathcal{D}'' \in \Lambda_S^2$, the $\mathcal{H}$-divergence between $\mathcal{D}'$ and $\mathcal{D}''$ is upper-bounded by the largest $\mathcal{H}$-divergence measured between elements of $S$.

\textbf{Lemma 1} \textit{(Bounding the $\mathcal{H}$-divergence between domains in the convex hull of the sources). Let $d_{\mathcal{H}}[\mathcal{D}^i_S, \mathcal{D}^k_S]\leq \epsilon, \; \forall \; i,k \in[N_S]$. The following inequality holds for the $\mathcal{H}$-divergence between any pair of domains $\mathcal{D}', \mathcal{D}'' \in \Lambda_S^2$}:
\begin{equation}
    d_{\mathcal{H}}[\mathcal{D}', \mathcal{D}''] \leq  \epsilon. 
\end{equation}

\textit{Proof.} C.f. Supplementary material. 

We thus argue that if one minimizes the maximum pairwise $\mathcal{H}$-divergence between source domains, which can be achieved by an encoding process that filters away domain discriminative cues, the $\mathcal{H}$-divergence between any two domains in $\Lambda_S$ also decreases. 

\subsection{Generalizing to unseen domains}
Now we turn our attention to the set of unseen distributions $\mathcal{D}^j_U$, $j\in[N_U]$, i.e., those in the support of the meta-distribution but not observed within the training sample. Given an unseen domain $\mathcal{D}_U$, we further introduce $\bar{\mathcal{D}}_U$, the element within $\Lambda_S$ which is closest to  $\mathcal{D}_U$, i.e.,  $\bar{\mathcal{D}}_U$ is given by $\argmin_{\pi_1, \ldots, \pi_{N_S}}d_{\mathcal{H}}\left[\mathcal{D}_U, \sum_{i=1}^{N_S} \pi_{i} \mathcal{D}^i_S\right]$. 
We now use Lemma 1 and previously proposed generalization bounds for the domain adaptation setting \cite{zhao2018adversarial, zhao2019learning} to derive a generalization bound for the risk $R_U[h]$. %in terms of $\epsilon$ and $d_{\mathcal{H}}[\bar{\mathcal{D}}_U, \mathcal{D}_U]$:

\textbf{Theorem 1} \textit{(Upper-bounding the risk on unseen domains). Given the previous setup, let $S$ be the set of source domains and  $\mathcal{Y} = [0, 1]$. The risk $R_U[h]$, $ \forall h \in \mathcal{H}$, for \textbf{any} unseen domain $\mathcal{D}_U$ such that $d_{\mathcal{H}}[\bar{\mathcal{D}}_U, \mathcal{D}_U] = \gamma$ is bounded as}:
\begin{equation}\label{eq:bound_1}
     R_U[h] \leq \sum_{i=1}^{N_S} \pi_{i} R^i_S[h] + \gamma+\epsilon +  \text{min}\{\mathbb{E}_{\bar{\mathcal{D}}_U}[|f_{S_{\pi}} - f_U|], \mathbb{E}_{\mathcal{D}_U}[|f_U - f_{S_\pi}|]\},     
\end{equation}
\textit{where $\epsilon$ is the highest pairwise $\mathcal{\tilde{H}}$-divergence measured between pairs within $S$, $\mathcal{\tilde{H}} = \{sign(|h(x) - h'(x)| - t) | \, h, h' \in \mathcal{H},0\leq t\leq1\}$ and $f_{S_{\pi}}(x) = \sum_{i=i}^{N_S} \pi_i f_{S_i}(x)$ is the labeling function for any $x\in \text{Supp}(\bar{\mathcal{D}}_U)$ resulting from combining all $f_{S_i}$ with weights $\pi_i$, $i\in[N_S]$, determined by $\bar{\mathcal{D}}_U$}.

\textit{Proof.} C.f. Supplementary material.

\textbf{Remark 1:} Notice that the right-most term in Theorem 1 accounts for the mismatch between the labeling functions $f_{S_{\pi}}$ and $f_U$, %and can also be written as $min\{R_{S_{\pi}}[f_U], R_{U}[f_{S_{\pi}}]\}$ in case we consider the 0-1 loss. 
which reduces to 0 in most adopted scenarios within domain adaption/generalization applications, since it is often considered that the \emph{covariate shift assumption} holds \cite{ben2010impossibility}. Under such setting, the labeling functions are the same across all domains in the support of $\mathfrak{D}$, i.e. $f_{S_i}=f_{U_j}=f$ for all $i\in [N_S]$ and $j\in[N_U]$. Besides the covariate shift assumption, previous work on multi-source domain adaptation \cite{hoffman2018algorithms} considered the case where the unseen domain $\mathcal{D}_U$ can be represented as a mixture of the sources with weights $\pi_i$, $i\in[N_S]$, i.e. $\mathcal{D}_U=\bar{\mathcal{D}}_U$. When such assumption holds, the term indicated by $\gamma$ in Theorem 1 will vanish. We thus re-state in Corollary 1 the previous result under such simplifying assumptions.

\textbf{Corollary 1} \textit{(Generalization to unseen domains within $\Lambda_S$ under the covariate shift assumption). Let all domains within the the support of the meta-distribution $\mathfrak{D}$ have labeling function $f$. Let $S$ be set of source domains and its convex-hull be denoted as $\Lambda_S$. The risk $R_U[h]$ of a hypothesis $h$ on an unseen domain $ \mathcal{D}_U \in \Lambda_S$, is upper-bounded by:}
\begin{equation}\label{eq:cor_1}
     R_U[h] \leq \sum_{i=1}^{N_S} \pi_{i} R^i_S[h] + \epsilon.
\end{equation}

\textit{Proof.} C.f. Supplementary material.

\textbf{Remark 2:} Based on the introduced results we define in the following an algorithm relying solely on source data, \emph{unlike domain adaptation approaches}. While the total source risk can be minimized as usual, $\epsilon$ can be minimized by encoding source data to a space where source domains are hard to distinguish. We remark that we empirically found (c.f. Sections \ref{exp:h_div} and \ref{exp:eeg}) the proposed algorithm was able to succeed even in scenarios where the considered assumptions are not likely to hold. 

\textbf{Remark 3:} We further highlight that the introduced results also provide insights regarding \emph{the importance of acquiring diverse datasets} in practice when targeting domain generalization (and hint as to why data augmentation is often helpful). The more diverse a dataset is regarding the number of domains present at training time, more likely it is that an unseen distribution lies within the convex hull of the source domains (i.e. $\gamma \rightarrow 0$). Therefore, not only the amount of data is important to achieve better generalization on unseen domains, but also the diversity of the training data is crucial. 

\textbf{Remark 4:} Another practical aspect worth remarking is that, even though our domain generalization setting is more general than ERM, Theorem 1 suggests that source domain labels should also be available, since they are required to estimate $\epsilon$, which is not the case for ERM. However, collecting domain labels is inherent to the data acquisition procedure for several tasks and commonly available as meta-data in cases such as, speech recognition, where different speakers or recording devices can be viewed as different domains.

\subsection{Practical contributions}
Motivated by the previous results, we propose to design algorithms that minimize the terms in the bound in (\ref{eq:bound_1}) that can be estimated even if only source data is observed, i.e., $\epsilon$ as well as the risks over the train sample. We thus aim at learning an encoder $E: \mathcal{X} \rightarrow \mathcal{Z}$, where $\mathcal{Z}\subset\mathbb{R}^d$ preserves information relevant for separating classes, while removing domain-specific cues in such a way that it is harder to distinguish examples from different domains in comparison to the original space $\mathcal{X}$.  

\textbf{Efficiently estimating $\boldsymbol\epsilon$:} Previous work on domain adaptation introduced strategies based on minimizing the empirical $\mathcal{H}$-divergence between sources and a given target domain \cite{ganin2016domain, zhao2018adversarial}. Instead, as per the discussion following Theorem 1, the domain generalization setting requires estimating pairwise $\mathcal{H}$-divergences across all available sources, not considering target data of any sort. Naively extending previous methods to our case would require $\mathcal{O}(N_S^2)$ estimators, which is unpractical given real-world cases where several source domains are available. We thus propose to use {\em one-vs-all} classifiers. In this case, there is one domain discriminator per source domain and the $k$-th discriminator estimates $\sum_{l \neq k} d_{\mathcal{H}}[\mathcal{D}^k_S, \mathcal{D}^l_S]$, and improves the method to a number of $\mathcal{H}$-divergence estimators linear on $N_S$.

\textbf{Training:} 
The proposed approach contains three main modules, all parameterized by neural networks: an encoder $E$ with parameters $\phi$, a task classifier $C$ with parameters $\theta_C$, and a set of $\mathcal{H}$-divergence estimators $D_k$ with parameters $\theta_{k}$, $k\in[N_S]$. Intuitively, $E$ attempts to minimize a classification loss $\mathcal{L}_C(\cdot;\theta_C)$ (standard cross-entropy in our case) and empirical $\mathcal{H}$-divergences, which is achieved through the maximization of domain discrimination losses, denominated $\mathcal{L}_k$. Each domain discriminator, on the other hand, aims at minimizing $\mathcal{L}_k$. The procedure for estimating $\phi$, $\theta_T$, and all $\theta_{k}$'s can be thus formulated as the following multiplayer minimax game:   
\begin{equation}
        \min_{\phi, \theta_C} \max_{\theta_1, \ldots, \theta_{N_S}} \mathcal{L}_C (C(E(x; \phi); \theta_C), y_C) - \sum_{k=1}^{N_S} \mathcal{L}_k(D_k(E(x; \phi); \theta_k), y_{k}),
\label{eq:game}
\end{equation}
% \begin{equation}
%     \min_{\phi, \theta_C} \max_{\theta_1, \ldots, \theta_{N_S}} \frac{1}{m} \sum_{y_{C_k}=1}^{N_C} y_{C_k}\text{log}(C_\theta(E_\phi(x)) - \sum_{k=1}^{N_S} \text{log}(\eta - \mathcal{L}_k(D_j(E(x; \phi); \theta_j), y_{j})),
% \end{equation}
where $y_{C}$ corresponds to the task label for the example $x$, and $y_k$ is equal to 1 in case $x\sim\mathcal{D}^k_S$, or 0 otherwise. 
Training is carried out with alternate updates. A pseudocode describing the training procedure is presented in Algorithm \ref{pseudo_code}. To further illustrate our proposed approach, we provide in Figure \ref{fig:method} (in the Supplementary material) a diagram showing the main components of the model in a case where three domains are available at training time.

\begin{algorithm*}[h]
\caption{Generalizing to unseen Domains via Distribution Matching}\label{pseudo_code}
\begin{algorithmic}[1]
\STATE Requires: classifier and encoder learning rate ($\beta_C$), domain discriminators learning rate ($\beta_D$), scaling ($\alpha$), mini-batch size ($m$).
\STATE Initialize $\phi$, $\theta_C$, $\theta_1, \ldots, \theta_{N_S}$ as $\phi^0$, $\theta_C^0$, $\theta_1^0, \ldots, \theta_{N_S}^0$. 
\FOR {$t = 1, \ldots, $ number of iterations}
\STATE Sample one mini-batch from each source domain $\{(x_1^i, y_C^i, y_1^i, \ldots, y_{N_S}^i)\}_{i=1}^{m}$
\STATE \texttt{$\#$ Update domain discriminators}
\FOR {$k = 1, \ldots, N_S$}
\STATE $\theta_k^{t} \gets \theta_k^{t-1} + \frac{\beta_D}{N_S \cdot m} \sum_{i=1}^{N_S \cdot m} \nabla_{\theta_k} \mathcal{L}_k(D_k(E(x^i; \phi^{t-1}); \theta_k^{t-1}), y_k^i)$
\ENDFOR
\STATE \texttt{$\#$ Update task classifier}
\STATE $\theta_C^{t}\gets \theta_C^{t-1} + \frac{\beta_C}{N_S \cdot m} \sum_{i=1}^{N_S \cdot m}\nabla_{\theta_C} \mathcal{L}_C (C(E(x^i; \phi^{t-1}); \theta_C^{t-1}), y_C^i)$
\STATE \texttt{$\#$ Update encoder}
\STATE $\phi^{t}\gets \phi^{t-1}$ $+ \frac{\beta_C}{N_S \cdot m} (\sum_{i=1}^{N_S \cdot m} \alpha \nabla_{\phi} \mathcal{L}_C (C(E(x^i; \phi^{t-1}); \theta_C^{t-1}), y_C^i)$ \\ 
$\; \qquad - (1-\alpha) \nabla_{\theta_k} \mathcal{L}_k(D_k(E(x^i; \phi^{t-1}); \theta_k^{t}), y_k^i))$
\ENDFOR
\end{algorithmic}
\end{algorithm*}

\textbf{Improving training stability:} Previous work on domain adaptation/generalization \cite{ganin2016domain,li2018deep} proposed solving the problem stated in (\ref{eq:game}) using a gradient reversal layer \cite{ganin2015unsupervised}. We empirically observed such approach to be heavily dependent on the choice of hyperparameters in order for training to converge. We propose to augment the described adversarial approach using strategies originally utilized for stabilizing the training of generative adversarial networks with multiple discriminators \cite{neyshabur2017stabilizing, albuquerque2019multi}. Namely, we include a random projection layer in the input of each domain discriminator with the goal of making examples from different distributions harder to be distinguished. In addition, we use the negative log-hypervolume instead of the summation in the game represented in (\ref{eq:game}) in order to assign more preference to solutions which decrease all pairwise divergences uniformly even in cases where there is a trade-off in their minimization. We refer to the proposed approach as G2DM (Generalizing to unseen Domains via Distribution Matching).

\textbf{Differences to multi-source domain adaptation:} We further remark the differences between G2DM and previous adversarial approaches which are often employed in domain adaptation. Essentially, G2DM compares examples \emph{only from source domains} to learn domain-agnostic representations, i.e., there is no notion of target distribution. Other settings such as \cite{sun2015survey, peng2019moment} are more restricted in that a particular distribution is targeted and data from that distribution is required, besides the source data we use in our case. Moreover, those approaches do not aim at matching source distributions and only consider $\mathcal{H}$-divergences computed between each source domain and the given target. In the case of G2DM, on the other hand, the goal is to match source domain distributions to decrease $\epsilon$, and thus only pairwise discrepancies between training domains are considered.

\section{Experimental Setup and Results}\label{sec:res}
% what happens when we leverage domain information and go beyond the i.i.d. setting?
% G2DM vs others
% Are h-div decreasing?
% Different stopping criterion
% Real-world experiments
% Supplementary material

We design our empirical evaluation to validate G2DM in conditions where different assumptions are satisfied. In the first scenario, we chose experimental conditions such that the covariate shift assumption holds. For that, we employ G2DM on object recognition tasks. In this case, we aim to answer the following research questions: i) Can G2DM perform better than standard ERM under i.i.d. assumptions by using information of source domains only? ii) Where does G2DM performance stand in comparison to previously proposed domain generalization strategies? iii) Is G2DM indeed enforcing distribution matching across source and unseen domains? And iv) What is the effect on the resulting performance given by different access models to test distributions during training?

We then evaluate whether G2DM is able to attain good out-of-domain performance even in the challenging scenario where the covariate shift assumption is likely to be violated. For that, we consider a real-world task that involves classifying electroencephalography (EEG) time series for affective state prediction, a burgeoning area within the human-machine systems field. In applications involving EEG data, subjects are often considered as distinct domains with different labeling functions \cite{albuquerque2019cross}. Such shift can be attributed to environmental and anatomic factors such as device placement and scalp characteristics \cite{wu2015reducing, wei2018subject}. %Additionally, in the supplementary material we provide extra results showing the impact of the random projection layer size and the number of source domains on final performance.

\subsection{Evaluation under covariate shift}
The VLCS benchmark \cite{fang2013unbiased} is composed by examples from five overlapping classes from the VOC2007 \cite{everingham2010pascal}, LabelMe \cite{russell2008labelme}, Caltech-101 \cite{griffin2007caltech}, and SUN \cite{choi2010exploiting} datasets. PACS \cite{li2017deeper}, in turn, consists of images distributed into seven classes from four different datasets: photo (P), art painting (A), cartoon (C), and sketch (S). We compare the performance of our proposed approach with a model trained with no mechanism to enforce domain generalization (referred to as ERM throughout this Section). Moreover, we consider the recently introduced invariant risk minimization (IRM) strategy \cite{arjovsky2019invariant} and include results reported in the literature achieved by Epi-FCR \cite{li2019episodic}, JiGen \cite{carlucci2019domain} along with the ERM results they provided (referred to as ERM-JiGen), and MMD-AAE \cite{li2018domain}. Finally, the adaptation of DANN for domain generalization reported in \cite{li2019episodic} was also considered. All such methods have as encoder the convolutional stack of AlexNet \cite{krizhevsky2012imagenet}. %and the weights are initialized from the pre-trained model on ImageNet \cite{deng2009imagenet}. 
Further implementation details can be found in the Supplementary material.

In Tables \ref{res:tab_vlcs} and \ref{res:tab_pacs}, we report the average best accuracy across three runs with different random seeds on the test partition of the unseen domain under a leave-one-domain-out validation scheme. Results show that G2DM outperforms ERM in terms of average performance across the unseen domains for both benchmarks, and supports the claim that leveraging source domain information as done by G2DM provides an improvement on generalization to unseen distributions in comparison to simply considering the i.i.d. requirement is satisfied. G2DM further presented better average performance when compared to our implementation of IRM, as well as results from other methods previously reported in the literature. We finally highlight that G2DM showed an improvement in performance in more challenging domains \cite{li2017deeper}, such as LabelMe and Sketch. 

\begin{minipage}{\textwidth}
\centering
\begin{minipage}[t]{0.49\textwidth}
\captionof{table}{Classification accuracy ($\%$) on VLCS datasets for models trained with leave-one-domain-out validation.}
\centering
\resizebox{\columnwidth}{!}{
\begin{tabular}{cccccc}
\hline
Unseen domain ($\rightarrow$)  & V      & L         & C                               & S                               & Average                         \\ \hline
DANN                                        & 66.40          & 64.00          & 92.60          & 63.60          & 71.70          \\
MMD-AAE                                     & 67.70          & 62.60          & 94.40          & 64.40          & 72.28          \\
Epi-FCR                                     & 67.10          & 64.30          & 94.10          & 65.90          & 72.90          \\
JiGen                                       & 70.62          & 60.90          & 96.93          & 64.30          & 73.19          \\
ERM - JiGen                                 & 71.96          & 59.18          & 96.93          & 62.57          & 72.66          \\
IRM     & 72.16    & 62.36    & 98.35 & 67.82  & 75.17                           \\
ERM     & 73.44 & 60.44 & 97.88    & 67.92                           & 74.92                           \\ \hline
G2DM    & 71.14    & 67.63 & 95.52                           & 69.37 & 75.92 \\ \hline
\end{tabular}
}
\label{res:tab_vlcs}
\end{minipage}
%\hspace{0.1cm}
\begin{minipage}[t]{0.49\textwidth}
\centering
\captionof{table}{Classification accuracy ($\%$) on PACS datasets for models trained with leave-one-domain-out validation.}
\resizebox{\columnwidth}{1.5cm}{
\begin{tabular}{cccccc}
\hline
Unseen domain ($\rightarrow$)     & P     & A     & C     & S     & Average \\ \hline

DANN      & 88.10  & 63.20  & 67.50  & 57.00    & 69.00      \\
Epi-FCR   & 86.10  & 64.70  & 72.30  & 65.00 & 72.00      \\
JiGen     & 89.00    & 67.63 & 71.71 & 65.18 & 73.38   \\
ERM - JiGen  & 89.98 & 66.68 & 69.41 & 60.02 & 71.52   \\
IRM        & 89.97 & 64.84 & 71.16  & 63.63 & 72.39         \\
ERM         & 90.02 & 64.86 & 70.18 & 61.40  & 71.61   \\ \hline
G2DM       & 88.12 & 66.60  & 73.36 & 
 66.19 & 73.55   \\ \hline
\end{tabular}
}
\label{res:tab_pacs}
\end{minipage}
\end{minipage}

\vspace{0.3cm}
\subsubsection{Checking $\boldsymbol{\mathcal{H}}$-divergences across sources and unseen domains}\label{exp:h_div}
\begin{figure}[h]
	\centering
	\subfloat[photo][Photo.]{\includegraphics[width=0.24\textwidth]{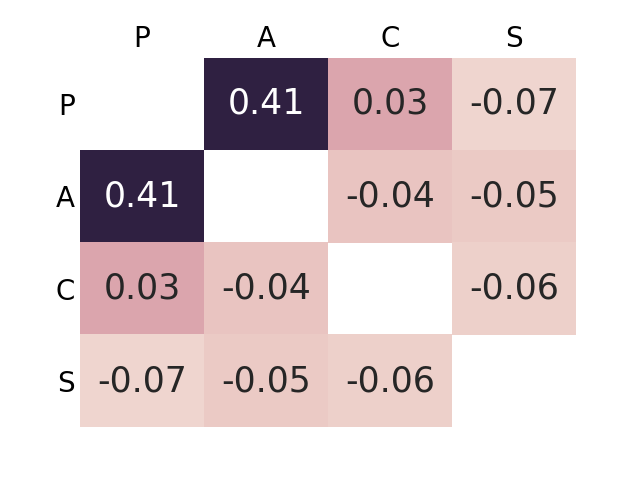}}
	\;
	\subfloat[art][Art.]{\includegraphics[width=0.24\textwidth]{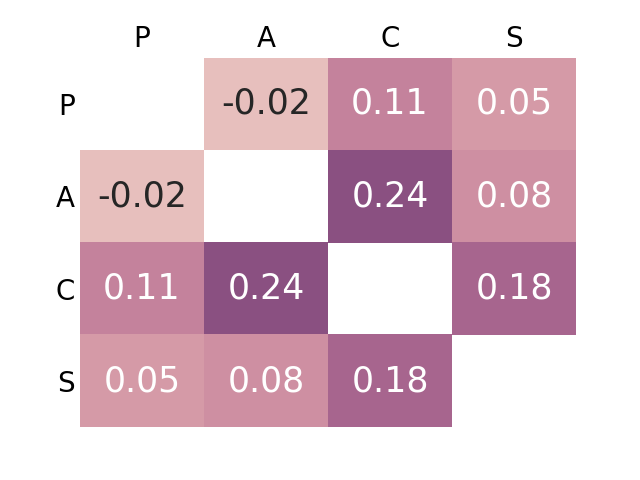}}
    \;
	\subfloat[cartoon][Cartoon.]{\includegraphics[width=0.24\textwidth]{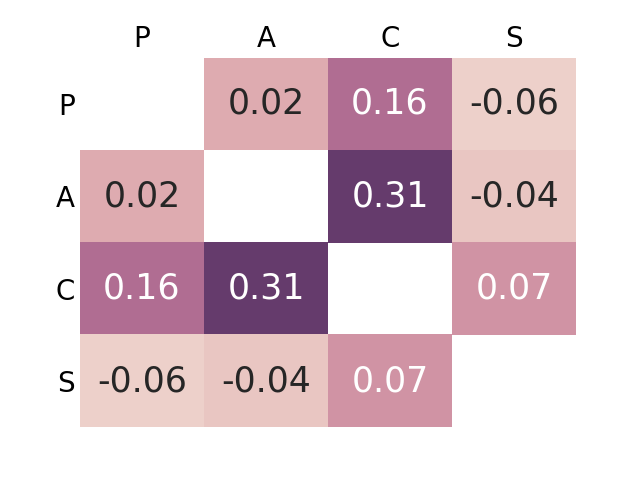}}
	\;
	\subfloat[sketch][Sketch.]{\includegraphics[width=0.24\textwidth]{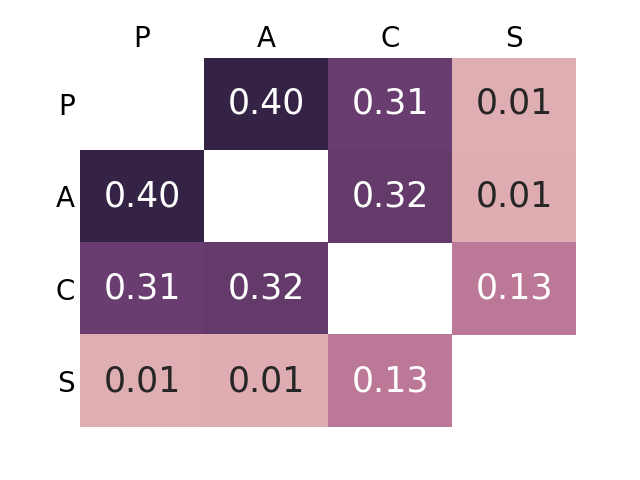}}
	\caption{Differences between estimated pairwise $\mathcal{H}$-divergences under ERM and G2DM on PACS (captions denote unseen domains). Higher values indicate that G2DM better matched domains. Overall, G2DM is able to decrease pairwise discrepancies.}
	\label{fig:a_dist}
 \end{figure}

We now investigate whether cross-domain $\mathcal{H}$-divergences are being in fact reduced by G2DM. We use ERM as a baseline as it does not include any mechanism to enforce distribution matching. We estimate $\mathcal{H}$-divergences by computing the proxy pairwise $\mathcal{A}$-distance \cite{ben2007analysis} for each pair of domains on the PACS benchmark. Classifiers are trained on top of the representations $\mathcal{Z}$ obtained with ERM and G2DM. We show in Figures \ref{fig:a_dist} the differences in estimated discrepancies between ERM and G2DM for each unseen domain. Each entry corresponds to a pair of domains indicated in the row and the column and positive values indicate that G2DM \emph{decreased} the corresponding pairwise $\mathcal{A}$-distance in comparison to ERM. Notice that the diagonals are left blank as we do not compute the classification accuracy between the same domains. 

We observe that, apart from the case where `photo' is the test domain, G2DM was in fact able to better match most of the source distributions, thus yielding smaller $\epsilon$ which favours generalization as predicted by Theorem 1. Notably, we highlight that although our proposed approach has no access to data from the unseen domain at training time and, therefore, does not directly implement a strategy to decrease the divergence between the unseen domain and the convex hull of the sources (i.e. $\gamma$), the results presented in Figure \ref{fig:a_dist} show that the estimated pairwise $\mathcal{H}$-divergence between the unseen domain and sources also decreased in most of the considered cases. 

In fact, the only mechanism the encoder has in order to reduce $\epsilon$ corresponds to learning how to filter domain information from the data, in the sense that once samples from two distinct distributions are encoded, one cannot distinguish from which distribution each sample came from. Observed results thus suggest that such encoder also removes domain information from the unseen distributions observed at test time, preventing the learning algorithm to yield a high $\gamma$.

 % Overall, G2DM is able to decrease the discrepancies between the unseen and source domains, as well as pairwise $\mathcal{H}$-divergences across source domains.
%, which preserves more domain information given that it is possible to distinguish data sources with higher accuracy in their case.%Further conclusions can be drawn from such results also show that when Photo is the unseen domain, our method did not need to find a representation which was invariant to Sketch, indicating that Art painting and Cartoon might already suffice to represent the Photo domain in the representation space.  

% It is hard to compare with previous work
% In the DG setting, the standard validation criteria might not be meaningful 
% We report results across different stopping criteria
% Results are not stable across different criteria
% People should report performance on different criteria and specify what was done

\subsubsection{The effect of different access methods to test data during training} 
Results of previous experiments correspond to an optimistic scenario where data from the unseen domain is made available for selecting the best performing model. This is not the case in practice since varying unseen distributions might appear. In Table~\ref{tab:stop_crit}, we compare results obtained further considering different access methods to the test data. Namely, we consider the case where no access to the unseen distribution is allowed and only source data can be used in order to define stopping criteria. In such cases, both validation accuracy and training loss computed on left-out in-domain data are employed (referred in Table~\ref{tab:stop_crit} as source accuracy and source loss, respectively). Moreover, as a reference of performance, we further report the accuracy achieved assuming access to unseen domain data during training in order to select the best model (referred in Table~\ref{tab:stop_crit} as unseen accuracy). For comparison, we further present the performance reported by \cite{li2018deep} for CIDDG, since a stopping criterion using solely data from source domains was employed in that case. We observe that, when using the training loss as stopping criterion, our strategy outperforms CIDDG for almost all domains, while the baseline performance severely degrades when `sketch' is the unseen domain. 

As an alternative to AlexNet, we further evaluate the performance of the proposed approach using the convolutional stack of a ResNet-18 \cite{he2016deep}, since it has shown promising results in recent work \cite{carlucci2019domain}. We compare our approach with JiGen\footnote{\label{fn:jigen}Results are generated using JiGen authors' source code (\url{https://github.com/fmcarlucci/JigenDG}).} adopting the same previously-discussed test data access methods for both approaches. We further report in Table~\ref{tab:stop_crit} the performance obtained by JiGen as reported in \cite{carlucci2019domain} although it is unclear which stopping criteria were adopted for that case. We observe that replacing AlexNet by ResNet-18 yields a more stable average performance across stopping criteria. Based on the results obtained with AlexNet, we remark that results might be too optimistic/pessimistic depending on the assumed access method to unseen distributions, and as such, in order to allow fair comparison between different approaches, \emph{the performance across different access methods should be reported}.

\begin{table}[h]
\centering
\caption{Accuracy ($\%$) on PACS with different stopping criteria.}
\resizebox{0.6\columnwidth}{!}{
\begin{tabular}{ccccccc}
\hline
  Method & Criterion  & P     & A     & C     & S & Average    \\ \hline
                \multicolumn{7}{c}{\textbf{AlexNet}}                                         \\ \hline
CIDDG \cite{li2018deep}                 & \footnotesize{From \cite{li2018deep}}  & 78.65 & 62.70 & 69.73 & 64.45 & 68.88          \\  \hdashline
\multirow{3}{*}{G2DM} & Source acc.  & 85.33 & 57.76 & 69.71 & 49.45 & 65.56     \\
                      & Source loss & 87.37 & 66.70 & 70.26 & 50.98 & 68.82   \\
                      & Unseen acc. & 88.80 & 66.70 & 73.29 & 65.03 & 73.45 \\ \hline
\multicolumn{7}{c}{\textbf{ResNet-18}}                                                       \\ \hline
\multirow{4}{*}{JiGen \cite{carlucci2019domain}} & Source acc.  & 95.83 & 78.52 & 73.31 & 69.14 & 79.20 \\ 
                       & Source loss & 95.83 & 78.89 & 73.32 & 70.73 & 79.69 \\
                       & Unseen acc. & 96.11 & 79.56 & 74.25 & 71.00 & 80.23 \\
                       & \footnotesize{From \cite{carlucci2019domain}} & 96.03 & 79.42 & 75.25 & 71.35 & 80.51 \\  \hdashline
\multirow{3}{*}{G2DM} & Source acc.  & 93.70 & 79.22 & 76.34 & 75.14 & 81.10   \\
                      & Source loss & 93.75 & 77.78 & 75.54 & 77.58 & 81.16 \\
                      & Unseen acc. & 94.63 & 81.44 & 79.35 & 79.52 & 83.34   \\
                       \hline  
%Epi-FCR                & \small{\cite{li2019episodic}} & 93.90 & 82.10 & 77.00 & 73.00 & 81.50 \\ \hline                       

\end{tabular}
}
\label{tab:stop_crit}
\end{table}

%One straightforward way to do so consists of selecting the best performing model on the validation partition of the source domains and then finally evaluate it on the target data. However, as previously pointed-out by \cite{d2018domain}, this quantity might not be informative in the domain generalization setting. We thus assess the performance of our proposed method considering two stopping criteria: i) validation accuracy on the source domains and ii) training task loss.  The same hyperparameters as for the results in Table \ref{res:tab_pacs} are used.

\subsection{Evaluation beyond the covariate shift assumption}\label{exp:eeg}
% Domain information comes for ''free'' during the data collection
% We get very close to the performance of models that were obtained for specific subjects with a general model
% We use less diverse training data in our protocol
% Validation loss contains different domains from the source
% Define semi and privileged baselines

%, namely DAN \cite{long2015learning}, DANN \cite{ganin2016domain}, MDAN \cite{zhao2018adversarial}, and MDMN \cite{li2018extracting}, as reported by \cite{li2018extracting}

We proceed to evaluate G2DM on a unfavorable scenario where the covariate shift is unlikely to hold. The goal of the selected task is to perform affective state estimation with three classes (positive, neutral, or negative) based on EEG signals from the SEED dataset \cite{zheng2015investigating} collected from 15 subjects. We use the architecture described in \cite{li2017targeting} for both G2DM and ERM. %We consider each subject as a different domain and perform leave-one-subject-out evaluation. 
For each subject left out for testing, we use 10 out of the remaining 14 domains for training and use the other 4 as validation data.

We report in Table~\ref{tab:seed_acc} the classification accuracy ($\%$) averaged across all unseen subjects and three independent training runs. Under \textbf{source data validation}, the reported performance was computed on the epoch of highest accuracy on the validation partition. The results under \textbf{semi-privileged} were obtained on the epoch of highest accuracy on the unseen subject data. The comparison between G2DM and ERM shows that even in this challenging case where the mismatch between labeling functions is not negligible, G2DM is able to successfully leverage the available domain information (which in this case comes with no additional effort at the data collection) and presents an improvement of more than $3.4\%$ in accuracy in comparison to ERM in both considered scenarios for the DG setting. 

\textbf{Comparison with domain adaptation strategies:} We further report in Table \ref{tab:seed_acc} results obtained by domain adaptation strategies (DA). Such methods, reported in Table \ref{tab:seed_acc} under \textbf{privileged baselines}, are privileged in the sense that unlabeled data belonging to the unseen domain (unknown in our case) is used to adapt representations at training time in order to yield \textbf{subject-specific} models. When comparing the DA strategies with our proposed domain generalization (DG) approach, we remark that DG strategies aim to obtain domain-agnostic models, as opposed to DA methods which target a specific distribution. As such, one would expect DA approaches to achieve better performance than DG. However, we observe G2DM's performance to be on par with, or even better than, some of the considered DA strategies. We conjecture a larger number of source domains available at training time would decrease the gap between DG and DA even further; i.e., it would be more likely that unseen domains are exactly represented in the convex hull of the sources yielding low $\gamma$ (c.f. Theorem 1).

% split the data of the 14 source subjects in train/validation splits regardless of the subjects, thus G2DM observed data from fewer subjects in its training set, (ii) it is not clear which stopping criterion whether the employed stopping criterion utilizes unseen domain labeled data. Adding this 
\begin{table}
\centering
\caption{Average accuracy ($\%$) on the SEED dataset across 15 subjects. Semi-privileged approaches correspond to the best performing model under the domain generalization setting. Privileged baselines (domain adaptation setting) have access to unseen domain data at training time.}
\resizebox{0.6\columnwidth}{!}{
\begin{tabular}{ccc}
\hline
                    Setting & Method & Average accuracy ($\%$)  \\ \hline 
\multirow{5}{*}{Domain generalization} & \multicolumn{2}{c}{\textit{Source data validation}} \\ \cline{2-3}
                    & ERM  & 51.98           \\
                    & G2DM & 55.77  \\ \cline{2-3}
                    & \multicolumn{2}{c}{\textit{Semi-privileged}} \\ \cline{2-3}
                    & ERM &  56.82  \\ 
                    & G2DM & 60.26  \\ \hline
\multicolumn{3}{c}{\textit{Privileged baselines}} \\ \hline
\multirow{4}{*}{Domain adaptation}   & DAN \cite{long2015learning,li2018extracting}  & 50.28                \\
                      & DANN \cite{ganin2016domain,li2018extracting} & 55.87                \\
                      & MDAN \cite{zhao2018adversarial,li2018extracting} & 56.65                \\
                      & MDMN \cite{li2018extracting} & 60.59                \\ \hline
\end{tabular}
}
\label{tab:seed_acc}
\end{table}

\section{Conclusion}\label{sec:conc}

% We tackled the domain generalization setting and showed generalization can be achieved in the neighborhood of the set of mixtures of distributions observed during training.

% New assumptions made over the data generating process can yield PAC type of results for domain complexity in a meta-distribution-agnostic fashion, which we intend to investigate in the future.

In this paper, we tackled the domain generalization problem and showed that generalization can be achieved in the neighborhood of the set of mixtures of distributions observed during training. Based on this result, we introduced G2DM, an efficient approach in yielding invariant representations across unseen distributions. Our method employs multiple one-vs-all domain discriminators, such that pairwise divergences between source distributions are estimated and minimized at training time. We provide empirical evidence supporting the claim that making use of domain information improves performance relative to standard settings relying on i.i.d. requirements. Moreover, the introduced approach outperformed recent methods which also leverage domain labels. We further showed that our proposed method resulted in strong results on a realistic setting, with performance comparable to privileged systems tailored to test distributions.

\bibliographystyle{IEEEtran}
\bibliography{references.bib}

\clearpage
\appendix
\section*{Supplementary Material}

\section{Proof of Lemma 1}
\textbf{Lemma 1.} \textit{Let $d_{\mathcal{H}}[\mathcal{D}^i_S, \mathcal{D}^k_S]\leq \epsilon, \; \forall \; i,k \in[N_S]$. The following inequality holds for the $\mathcal{H}$-divergence between any pair of domains $\mathcal{D}', \mathcal{D}'' \in \Lambda_S^2$}:

\begin{equation}
    d_{\mathcal{H}}[\mathcal{D}', \mathcal{D}''] \leq  \epsilon. 
\end{equation}

\textit{Proof.} Consider two unseen domains, $\mathcal{D}'_U$ and $\mathcal{D}''_U$ on the convex-hull $\Lambda_S$ of $N_S$ source domains with support $\Omega$. Consider also $\mathcal{D}'_U(\cdot) = \sum_{k=1}^{N_S} \pi_{k} \mathcal{D}_S^k(\cdot)$ and $\mathcal{D}''_U(\cdot) = \sum_{l=1}^{N_S} \pi_{l} \mathcal{D}_S^l(\cdot)$
The $\mathcal{H}$-divergence between $\mathcal{D}'_U$ and $\mathcal{D}''_U$ can be written as:

\begin{equation}
\begin{split}
 d_{\mathcal{H}}[\mathcal{D}'_U, \mathcal{D}''_U] =& 2 \sup_{h \in \mathcal{H}} | \text{Pr}_{x\sim\mathcal{D}'_U} [h(x)=1]  - \text{Pr}_{x\sim\mathcal{D}''_U} [h(x)=1] |,  \\
=& 2 \sup_{h \in \mathcal{H}} | \mathbb{E}_{x\sim\mathcal{D}'_U} [\mathbf{I}(h(x))]  - \mathbb{E}_{x\sim\mathcal{D}''_U} [\mathbf{I}(h(x))] |, \\    
=& 2 \sup_{h \in \mathcal{H}} \left| \int_{\Omega} \mathcal{D}'_U(x) \mathbf{I}(h(x)) dx  - \int_{\Omega} \mathcal{D}''_U(x) \mathbf{I}(h(x)) dx \right|, \\    
=& 2 \sup_{h \in \mathcal{H}} \left| \int_{\Omega} \sum_{k=1}^{N_S} \pi_{k} \mathcal{D}^k_S(x)  \mathbf{I}(h(x)) dx  - \int_{\Omega} \sum_{l=1}^{N_S} \pi_{l} \mathcal{D}^l_S(x) \mathbf{I}(h(x)) dx \right|, \\    
=& 2 \sup_{h \in \mathcal{H}} \left| \int_{\Omega} \sum_{l=1}^{N_S} \sum_{k=1}^{N_S} \pi_{l} \pi_{k} \mathcal{D}^k_S(x)  \mathbf{I}(h(x)) dx  - \int_{\Omega} \sum_{l=1}^{N_S} \sum_{k=1}^{N_S} \pi_{l} \pi_{k} \mathcal{D}^l_S(x) \mathbf{I}(h(x)) dx \right|, \\     
=& 2 \sup_{h \in \mathcal{H}} \left| \sum_{l=1}^{N_S} \sum_{k=1}^{N_S} \pi_{l} \pi_{k} \left( \int_{\Omega} \mathcal{D}^k_S(x)  \mathbf{I}(h(x)) dx  - \int_{\Omega}  \mathcal{D}^l_S(x) \mathbf{I}(h(x)) dx \right) \right|.
\end{split}
\end{equation}

Using the triangle inequality, we can write:
\begin{equation}
 d_{\mathcal{H}}[\mathcal{D}'_U, \mathcal{D}''_U] \leq 2 \sup_{h \in \mathcal{H}} \sum_{l=1}^{N_S} \sum_{k=1}^{N_S} \pi_{l} \pi_{k} \left| \int_{\Omega} \mathcal{D}^k_S(x)  \mathbf{I}(h(x)) dx  - \int_{\Omega}  \mathcal{D}^l_S(x) \mathbf{I}(h(x)) dx \right|.    
\end{equation}

Finally, using the sub-additivity of the $\sup$:
\begin{equation}
\begin{split}
d_{\mathcal{H}}[\mathcal{D}'_U, \mathcal{D}''_U] \leq &\sum_{l=1}^{N_S} \sum_{k=1}^{N_S} \pi_{l} \pi_{k} 2 \sup_{h \in \mathcal{H}} \left| \int_{\Omega} \mathcal{D}^k_S(x)  \mathbf{I}(h(x)) dx  - \int_{\Omega} \mathcal{D}^l_S(x) \mathbf{I}(h(x)) dx \right|, \\
= &\sum_{l=1}^{N_S} \sum_{k=1}^{N_S} \pi_{l} \pi_{k}  d_{\mathcal{H}}[\mathcal{D}^k_S, \mathcal{D}^l_S]. 
\end{split}
\end{equation}

Given $d_{\mathcal{H}}[\mathcal{D}^k_S, \mathcal{D}^l_S] \leq \epsilon$ $\forall \; k,l \in[N_S]$:
\begin{equation*}
d_{\mathcal{H}}[\mathcal{D}'_U, \mathcal{D}''_U] \leq \epsilon. \qquad \qquad \square   
\end{equation*}

\section{Proof of Theorem 1}
\textbf{Theorem 1.} \textit{Let $S$ be the set of source domains and  $\mathcal{Y} = [0, 1]$. The risk $R_U[h]$, $ \forall h \in \mathcal{H}$, for \textbf{any} unseen domain $\mathcal{D}_U$ such that $d_{\mathcal{H}}[\bar{\mathcal{D}}_U, \mathcal{D}_U] = \gamma$, is bounded as:}
\begin{equation}\label{eq:bound_1}
     R_U[h] \leq \sum_{i=1}^{N_S} \pi_{i} R^i_S[h] + \gamma+\epsilon +
     \text{min}\{\mathbb{E}_{\bar{\mathcal{D}}_U}[|f_{S_{\pi}} - f_U|], \mathbb{E}_{\mathcal{D}_U}[|f_U - f_{S_\pi}|]\},
\end{equation}
\textit{where $\epsilon$ is the highest pairwise $\mathcal{\tilde{H}}$-divergence measured between pairs of domains within $S$, $\mathcal{\tilde{H}} = \{sign(|h(x) - h'(x)| - t) | h, h' \in \mathcal{H},0\leq t\leq1\}$ and $f_{S_{\pi}}(x) = \sum_{i=i}^{N_S} \pi_i f_{S_i}(x)$ is the labeling function for any $x\in \text{Supp}(\bar{\mathcal{D}}_U)$ resulting from combining all $f_{S_i}$ with weights $\pi_i$, $i\in[N_S]$, determined by $\bar{\mathcal{D}}_U$}.

\textit{Proof of Theorem 1.} Let the source and target domains be $\langle \mathcal{D}_S, f_S \rangle$ and $\langle \mathcal{D}_T, f_T \rangle$, respectively. For the single-source, single-target domain adaptation case, it was previously shown that the risk of any $h \in \mathcal{H}$, $h: \mathcal{X} \rightarrow [0, 1]$ is bounded by \cite{zhao2019learning}:
\begin{equation}\label{eq:zhao_19}
    R_T[h] \leq R_S[h] + d_{\mathcal{\tilde{H}}}[\mathcal{D}_S, \mathcal{D}_T]
    + \min \{\mathbb{E}_{\mathcal{D}_S}[|f_S - f_T|],
    \mathbb{E}_{\mathcal{D}_T}[|f_T - f_S|]\},    
\end{equation}
where $\mathcal{\tilde{H}} = \{sign(|h(x) - h'(x)| - t) | h, h' \in \mathcal{H},0\leq t\leq1\}$.

In order to devise a generalization bound for the risk on any unseen domain in terms of quantities related to the distributions seen at training time, we start by writing (\ref{eq:zhao_19}) considering $\mathcal{D}_U$ and its ``projection'' onto the convex-hull of the sources $\bar{\mathcal{D}}_U=\argmin_{\pi_1, \ldots, \pi_{N_S}}d_{\mathcal{H}}\left[\mathcal{D}_U, \sum_{i=1}^{N_S} \pi_{i} \mathcal{D}^i_S\right]$. For that, we introduce the labeling function $f_{S_{\pi}}(x) = \sum_{i=i}^{N_S} \pi_i f_{S_i}(x)$, which is an ensemble of the respective labeling functions from each source domain weighted by the mixture coefficients that determine $\bar{\mathcal{D}_U}$. $R_U[h]$ can thus be bounded as:
\begin{equation*}\label{eq:bound_12}
        R_U[h] \leq R_{\bar{U}}[h] + d_{\mathcal{\tilde{H}}}[\bar{\mathcal{D}}_U, \mathcal{D}_U] + \min \{\mathbb{E}_{\bar{\mathcal{D}}_{U}}[|f_{S_{\pi}} - f_U|],
    \mathbb{E}_{\mathcal{D}_U}[|f_U - f_{S_{\pi}}|]\}.
\end{equation*}

Similarly to the proof of our Lemma 1 for the case where $\mathcal{D}'=\mathcal{D}_U$ and $\mathcal{D}''=\bar{\mathcal{D}}_U$ (and to \cite{zhao2018adversarial}), it follows that:
\begin{equation}\label{eq:bound_13}
    R_U[h] \leq \sum_{i=1}^{N_S} \pi_i R^i_{S}[h] + \sum_{i=1}^{N_S} \pi_i d_{\mathcal{\tilde{H}}}[\mathcal{D}^i_{S}, \mathcal{D}_U]
    + \min \{\mathbb{E}_{\bar{\mathcal{D}}_{U}}[|f_{S_{\pi}} - f_U|],
    \mathbb{E}_{\mathcal{D}_U}[|f_U - f_{S_{\pi}}|]\}.
\end{equation}

Using the triangle inequality for the $\mathcal{H}$-divergence along with Lemma 1, we can bound the $\mathcal{\tilde{H}}$-divergence between $\mathcal{D}_U$ and any source domain $\mathcal{D}_S^i$, $d_{\mathcal{\tilde{H}}}[\mathcal{D}_U, \mathcal{D}_S^i]$, according to:

\begin{equation*}\label{eq:triangle}
\begin{split}
    d_{\mathcal{\tilde{H}}}[\mathcal{D}_U, \mathcal{D}_S^i] &\leq d_{\mathcal{\tilde{H}}}[\mathcal{D}_U, \bar{\mathcal{D}}_U] + d_{\mathcal{\tilde{H}}}[\bar{\mathcal{D}}_U, \mathcal{D}_S^i], \\
    &\leq \gamma + \epsilon,
\end{split}    
\end{equation*}
where $\gamma = d_{\mathcal{\tilde{H}}}[\mathcal{D}_U, \bar{\mathcal{D}}_U]$. Using this result, we can now upper-bound $\sum_{i=1}^{N_S} \pi_{i} d_{\mathcal{\tilde{H}}}[\mathcal{D}_U, \mathcal{D}^i_S]$ by $\gamma+\epsilon$ and finally re-write (\ref{eq:bound_13}) as:
\begin{equation*}\label{eq:bound_final}
 R_U[h] \leq \sum_{i=1}^{N_S} \pi_{i} R^i_S[h] + \gamma+\epsilon
 + \text{min}\{\mathbb{E}_{\bar{\mathcal{D}}_U}[|f_{S_{\pi}} - f_U|], \mathbb{E}_{\mathcal{D}_U}[|f_U - f_{S_\pi}|]\}.   \qquad \square
\end{equation*}

\section{Proof of Corollary 1}
\textbf{Corollary 1.} \textit{Let all domains within the the support of the meta-distribution $\mathfrak{D}$ have labeling function $f$. Let $S$ be set of source domains and its convex-hull be denoted as $\Lambda_S$. The risk $R_U[h]$ of a hypothesis $h$ on an unseen domain $ \mathcal{D}_U \in \Lambda_S$, is upper-bounded by:}
\begin{equation}\label{eq:cor_1}
     R_U[h] \leq \sum_{i=1}^{N_S} \pi_{i} R^i_S[h] + \epsilon.
\end{equation}
\textit{Proof of Corollary 1.} The right-most term of (\ref{eq:bound_1}) accounts for the mismatch between the labeling functions of $\mathcal{D}_U$ and $\bar{\mathcal{D}}_U$. Since all domains within $\mathfrak{D}$ have the same labeling function, this term is equal to 0. As $\mathcal{D}_U \in \Lambda_S$, $\mathcal{D}_U=\bar{\mathcal{D}}_U$, which results in $d_{\tilde{\mathcal{H}}}[\mathcal{D}_U, \bar{\mathcal{D}}_U]=\gamma=0$. $\square$

\section{One-vs-all $\mathcal{H}$-divergence estimation}
We illustrate the estimation of $\mathcal{H}$-divergences using one-vs-all discriminators by considering an example in which 3 source domains are available. Consider samples of size $M$ from $N_S=3$ source domains which are available at training time. The loss $\mathcal{L}_1$ for the domain discriminator $D_1$ accounting for estimating $d_{\mathcal{H}}[\mathcal{D}_1, \mathcal{D}_2]$ and $d_{\mathcal{H}}[\mathcal{D}_1, \mathcal{D}_3]$ can be written as:

\begin{equation}
\begin{split}
\mathcal{L}_1 &= \frac{1}{3M} \sum_{i=1}^{3M} \ell(D_1(x_i), y_1), \\
              &= \frac{1}{M} \sum_{i=1}^{M} \ell(D_1(x_i), y_1) + \frac{1}{M} \sum_{i=M+1}^{2M} \ell(D_1(x_i), y_1) + \frac{1}{M} \sum_{i=2M+1}^{3M} \ell(D_1(x_i), y_1),
\end{split}
\end{equation}
 
where $\ell$ represents a loss function (e.g. 0-1 loss) and each term accounts for the loss provided by examples from one domain. Splitting the first term in two parts and replacing the domain labels $y_1$ by their corresponding values, we obtain:

\begin{equation}\label{eq:hdiv_decomp}
\begin{split}
\mathcal{L}_1 &= \frac{1}{M} \sum_{i=1}^{M/2} \ell(D_1(x_i), 1) + \frac{1}{M} \sum_{i=M+1}^{2M} \ell(D_1(x_i), 0) \\ 
             &+ \frac{1}{M} \sum_{i=\frac{M}{2}+1}^{M} \ell(D_1(x_i), 1) + \frac{1}{M} \sum_{i=2M+1}^{3M} \ell(D_1(x_i), 0).
\end{split}
\end{equation}
The first two terms from Eq.\ref{eq:hdiv_decomp} account for $d_{\mathcal{H}}[\mathcal{D}_1, \mathcal{D}_2]$ and the last two terms account for $d_{\mathcal{H}}[\mathcal{D}_1, \mathcal{D}_3]$.

\section{Illustration}
\tikzset{every picture/.style={line width=0.75pt}} %set default line width to 0.75pt        
\begin{figure}[h]
\centering
\resizebox{0.4\columnwidth}{5cm}{

\begin{tikzpicture}[x=0.75pt,y=0.75pt,yscale=-1,xscale=1]

%Rounded Rect [id:dp5166036673169321] 
\draw  [fill={rgb, 255:red, 151; green, 221; blue, 204 }  ,fill opacity=1 ] (127,148) .. controls (127,143.58) and (130.58,140) .. (135,140) -- (210,140) .. controls (214.42,140) and (218,143.58) .. (218,148) -- (218,172) .. controls (218,176.42) and (214.42,180) .. (210,180) -- (135,180) .. controls (130.58,180) and (127,176.42) .. (127,172) -- cycle ;
%Rounded Rect [id:dp5874429076007261] 
\draw  [fill={rgb, 255:red, 151; green, 221; blue, 204 }  ,fill opacity=1 ] (135,158) .. controls (135,153.58) and (138.58,150) .. (143,150) -- (218,150) .. controls (222.42,150) and (226,153.58) .. (226,158) -- (226,182) .. controls (226,186.42) and (222.42,190) .. (218,190) -- (143,190) .. controls (138.58,190) and (135,186.42) .. (135,182) -- cycle ;
%Rounded Rect [id:dp9754641258440724] 
\draw  [fill={rgb, 255:red, 241; green, 161; blue, 226 }  ,fill opacity=1 ] (276,229.2) .. controls (276,224.67) and (279.67,221) .. (284.2,221) -- (380.8,221) .. controls (385.33,221) and (389,224.67) .. (389,229.2) -- (389,253.8) .. controls (389,258.33) and (385.33,262) .. (380.8,262) -- (284.2,262) .. controls (279.67,262) and (276,258.33) .. (276,253.8) -- cycle ;
%Rounded Rect [id:dp10852022153501828] 
\draw  [fill={rgb, 255:red, 158; green, 201; blue, 250 }  ,fill opacity=1 ] (285,59) .. controls (285,54.58) and (288.58,51) .. (293,51) -- (378,51) .. controls (382.42,51) and (386,54.58) .. (386,59) -- (386,83) .. controls (386,87.42) and (382.42,91) .. (378,91) -- (293,91) .. controls (288.58,91) and (285,87.42) .. (285,83) -- cycle ;
%Image [id:dp1997564921147601] 
\draw (60,119) node  {\includegraphics[width=39pt,height=36pt]{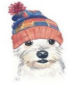}};
%Image [id:dp22379057145197878] 
\draw (56.5,185.5) node  {\includegraphics[width=33.75pt,height=30.75pt]{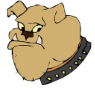}};
%Image [id:dp06173078577187585] 
\draw (57.5,251) node  {\includegraphics[width=39.75pt,height=34.5pt]{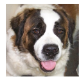}};
%Image [id:dp7577188307666647] 
\draw (495,244) node  {\includegraphics[width=39pt,height=36pt]{art.png}};
%Image [id:dp6699819941206819] 
\draw (628.5,241.5) node  {\includegraphics[width=33.75pt,height=30.75pt]{cartoon.png}};
%Image [id:dp5773839921266004] 
\draw (565.5,240) node  {\includegraphics[width=39.75pt,height=34.5pt]{photo.png}};
%Shape: Rectangle [id:dp03198351733693405] 
\draw  [color={rgb, 255:red, 144; green, 19; blue, 254 }  ,draw opacity=1 ][dash pattern={on 5.63pt off 4.5pt}][line width=1.5]  (460,215) -- (524,215) -- (524,269) -- (460,269) -- cycle ;
%Shape: Rectangle [id:dp32958142736967044] 
\draw  [color={rgb, 255:red, 208; green, 2; blue, 27 }  ,draw opacity=1 ][dash pattern={on 1.69pt off 2.76pt}][line width=1.5]  (534,215) -- (661,215) -- (661,269) -- (534,269) -- cycle ;
%Image [id:dp08028328612689783] 
\draw (626,349) node  {\includegraphics[width=39pt,height=36pt]{art.png}};
%Image [id:dp12719257028253428] 
\draw (494.5,351.5) node  {\includegraphics[width=33.75pt,height=30.75pt]{cartoon.png}};
%Image [id:dp46924999921268995] 
\draw (566.5,348) node  {\includegraphics[width=39.75pt,height=34.5pt]{photo.png}};
%Shape: Rectangle [id:dp037125148429036026] 
\draw  [color={rgb, 255:red, 144; green, 19; blue, 254 }  ,draw opacity=1 ][dash pattern={on 5.63pt off 4.5pt}][line width=1.5]  (461,323) -- (525,323) -- (525,377) -- (461,377) -- cycle ;
%Shape: Rectangle [id:dp9175016410892443] 
\draw  [color={rgb, 255:red, 208; green, 2; blue, 27 }  ,draw opacity=1 ][dash pattern={on 1.69pt off 2.76pt}][line width=1.5]  (535,323) -- (662,323) -- (662,377) -- (535,377) -- cycle ;
%Image [id:dp6218617829625566] 
\draw (573,457) node  {\includegraphics[width=39pt,height=36pt]{art.png}};
%Image [id:dp6462781463159695] 
\draw (626.5,458.5) node  {\includegraphics[width=33.75pt,height=30.75pt]{cartoon.png}};
%Image [id:dp7594259940843351] 
\draw (494.5,458) node  {\includegraphics[width=39.75pt,height=34.5pt]{photo.png}};
%Shape: Rectangle [id:dp44189323019981264] 
\draw  [color={rgb, 255:red, 144; green, 19; blue, 254 }  ,draw opacity=1 ][dash pattern={on 5.63pt off 4.5pt}][line width=1.5]  (462,431) -- (526,431) -- (526,485) -- (462,485) -- cycle ;
%Shape: Rectangle [id:dp7242963361158161] 
\draw  [color={rgb, 255:red, 208; green, 2; blue, 27 }  ,draw opacity=1 ][dash pattern={on 1.69pt off 2.76pt}][line width=1.5]  (536,431) -- (663,431) -- (663,485) -- (536,485) -- cycle ;
%Straight Lines [id:da9663920503608221] 
\draw [color={rgb, 255:red, 109; green, 109; blue, 109 }  ,draw opacity=1 ]   (98,165) -- (124,165) ;
\draw [shift={(126,165)}, rotate = 180] [color={rgb, 255:red, 109; green, 109; blue, 109 }  ,draw opacity=1 ][line width=0.75]    (10.93,-3.29) .. controls (6.95,-1.4) and (3.31,-0.3) .. (0,0) .. controls (3.31,0.3) and (6.95,1.4) .. (10.93,3.29)   ;

%Straight Lines [id:da46697287707911883] 
\draw [color={rgb, 255:red, 109; green, 109; blue, 109 }  ,draw opacity=1 ]   (233,180) -- (274.82,237.38) ;
\draw [shift={(276,239)}, rotate = 233.91] [color={rgb, 255:red, 109; green, 109; blue, 109 }  ,draw opacity=1 ][line width=0.75]    (10.93,-3.29) .. controls (6.95,-1.4) and (3.31,-0.3) .. (0,0) .. controls (3.31,0.3) and (6.95,1.4) .. (10.93,3.29)   ;

%Straight Lines [id:da07375842772215324] 
\draw [color={rgb, 255:red, 109; green, 109; blue, 109 }  ,draw opacity=1 ]   (233,180) -- (275.51,348.06) ;
\draw [shift={(276,350)}, rotate = 255.81] [color={rgb, 255:red, 109; green, 109; blue, 109 }  ,draw opacity=1 ][line width=0.75]    (10.93,-3.29) .. controls (6.95,-1.4) and (3.31,-0.3) .. (0,0) .. controls (3.31,0.3) and (6.95,1.4) .. (10.93,3.29)   ;

%Straight Lines [id:da3411023770283481] 
\draw [color={rgb, 255:red, 109; green, 109; blue, 109 }  ,draw opacity=1 ]   (233,180) -- (273.71,455.02) ;
\draw [shift={(274,457)}, rotate = 261.58] [color={rgb, 255:red, 109; green, 109; blue, 109 }  ,draw opacity=1 ][line width=0.75]    (10.93,-3.29) .. controls (6.95,-1.4) and (3.31,-0.3) .. (0,0) .. controls (3.31,0.3) and (6.95,1.4) .. (10.93,3.29)   ;

%Straight Lines [id:da3510019449622739] 
\draw [color={rgb, 255:red, 109; green, 109; blue, 109 }  ,draw opacity=1 ]   (233,180) -- (283.13,76.8) ;
\draw [shift={(284,75)}, rotate = 475.91] [color={rgb, 255:red, 109; green, 109; blue, 109 }  ,draw opacity=1 ][line width=0.75]    (10.93,-3.29) .. controls (6.95,-1.4) and (3.31,-0.3) .. (0,0) .. controls (3.31,0.3) and (6.95,1.4) .. (10.93,3.29)   ;

%Straight Lines [id:da2185601495771885] 
\draw [color={rgb, 255:red, 109; green, 109; blue, 109 }  ,draw opacity=1 ]   (386.5,71.5) -- (443.67,71.98) ;
\draw [shift={(445.67,72)}, rotate = 180.48] [color={rgb, 255:red, 109; green, 109; blue, 109 }  ,draw opacity=1 ][line width=0.75]    (10.93,-3.29) .. controls (6.95,-1.4) and (3.31,-0.3) .. (0,0) .. controls (3.31,0.3) and (6.95,1.4) .. (10.93,3.29)   ;

%Straight Lines [id:da7772840720734662] 
\draw [color={rgb, 255:red, 109; green, 109; blue, 109 }  ,draw opacity=1 ]   (389,242) -- (443.5,241.76) ;
\draw [shift={(445.5,241.75)}, rotate = 539.75] [color={rgb, 255:red, 109; green, 109; blue, 109 }  ,draw opacity=1 ][line width=0.75]    (10.93,-3.29) .. controls (6.95,-1.4) and (3.31,-0.3) .. (0,0) .. controls (3.31,0.3) and (6.95,1.4) .. (10.93,3.29)   ;

%Rounded Rect [id:dp021933066679070246] 
\draw  [color={rgb, 255:red, 155; green, 155; blue, 155 }  ,draw opacity=1 ] (446,207) .. controls (446,197.06) and (454.06,189) .. (464,189) -- (652,189) .. controls (661.94,189) and (670,197.06) .. (670,207) -- (670,261) .. controls (670,270.94) and (661.94,279) .. (652,279) -- (464,279) .. controls (454.06,279) and (446,270.94) .. (446,261) -- cycle ;
%Rounded Rect [id:dp4953155293252496] 
\draw  [color={rgb, 255:red, 155; green, 155; blue, 155 }  ,draw opacity=1 ] (446,61.2) .. controls (446,56.67) and (449.67,53) .. (454.2,53) -- (658.8,53) .. controls (663.33,53) and (667,56.67) .. (667,61.2) -- (667,85.8) .. controls (667,90.33) and (663.33,94) .. (658.8,94) -- (454.2,94) .. controls (449.67,94) and (446,90.33) .. (446,85.8) -- cycle ;
%Rounded Rect [id:dp5748950638230734] 
\draw  [color={rgb, 255:red, 155; green, 155; blue, 155 }  ,draw opacity=1 ] (447,313) .. controls (447,303.06) and (455.06,295) .. (465,295) -- (653,295) .. controls (662.94,295) and (671,303.06) .. (671,313) -- (671,367) .. controls (671,376.94) and (662.94,385) .. (653,385) -- (465,385) .. controls (455.06,385) and (447,376.94) .. (447,367) -- cycle ;
%Straight Lines [id:da9149074771730652] 
\draw [color={rgb, 255:red, 109; green, 109; blue, 109 }  ,draw opacity=1 ]   (389.4,345.4) -- (445.5,345.74) ;
\draw [shift={(447.5,345.75)}, rotate = 180.35] [color={rgb, 255:red, 109; green, 109; blue, 109 }  ,draw opacity=1 ][line width=0.75]    (10.93,-3.29) .. controls (6.95,-1.4) and (3.31,-0.3) .. (0,0) .. controls (3.31,0.3) and (6.95,1.4) .. (10.93,3.29)   ;

%Rounded Rect [id:dp3855740644409993] 
\draw  [color={rgb, 255:red, 155; green, 155; blue, 155 }  ,draw opacity=1 ] (447,422) .. controls (447,412.06) and (455.06,404) .. (465,404) -- (653,404) .. controls (662.94,404) and (671,412.06) .. (671,422) -- (671,476) .. controls (671,485.94) and (662.94,494) .. (653,494) -- (465,494) .. controls (455.06,494) and (447,485.94) .. (447,476) -- cycle ;
%Straight Lines [id:da8199808836552551] 
\draw [color={rgb, 255:red, 109; green, 109; blue, 109 }  ,draw opacity=1 ]   (389.8,452.6) -- (445,452.6) ;
\draw [shift={(447,452.6)}, rotate = 180] [color={rgb, 255:red, 109; green, 109; blue, 109 }  ,draw opacity=1 ][line width=0.75]    (10.93,-3.29) .. controls (6.95,-1.4) and (3.31,-0.3) .. (0,0) .. controls (3.31,0.3) and (6.95,1.4) .. (10.93,3.29)   ;

%Rounded Rect [id:dp8469389007319574] 
\draw  [color={rgb, 255:red, 155; green, 155; blue, 155 }  ,draw opacity=1 ] (19,84.4) .. controls (19,75.89) and (25.89,69) .. (34.4,69) -- (80.6,69) .. controls (89.11,69) and (96,75.89) .. (96,84.4) -- (96,268.6) .. controls (96,277.11) and (89.11,284) .. (80.6,284) -- (34.4,284) .. controls (25.89,284) and (19,277.11) .. (19,268.6) -- cycle ;
%Rounded Rect [id:dp5920831816476588] 
\draw  [fill={rgb, 255:red, 151; green, 221; blue, 204 }  ,fill opacity=1 ] (142,168) .. controls (142,163.58) and (145.58,160) .. (150,160) -- (225,160) .. controls (229.42,160) and (233,163.58) .. (233,168) -- (233,192) .. controls (233,196.42) and (229.42,200) .. (225,200) -- (150,200) .. controls (145.58,200) and (142,196.42) .. (142,192) -- cycle ;
%Rounded Rect [id:dp4735490679958205] 
\draw  [fill={rgb, 255:red, 241; green, 161; blue, 226 }  ,fill opacity=1 ] (277,334.2) .. controls (277,329.67) and (280.67,326) .. (285.2,326) -- (381.8,326) .. controls (386.33,326) and (390,329.67) .. (390,334.2) -- (390,358.8) .. controls (390,363.33) and (386.33,367) .. (381.8,367) -- (285.2,367) .. controls (280.67,367) and (277,363.33) .. (277,358.8) -- cycle ;
%Rounded Rect [id:dp7848957657431777] 
\draw  [fill={rgb, 255:red, 241; green, 161; blue, 226 }  ,fill opacity=1 ] (276,441.2) .. controls (276,436.67) and (279.67,433) .. (284.2,433) -- (380.8,433) .. controls (385.33,433) and (389,436.67) .. (389,441.2) -- (389,465.8) .. controls (389,470.33) and (385.33,474) .. (380.8,474) -- (284.2,474) .. controls (279.67,474) and (276,470.33) .. (276,465.8) -- cycle ;
%Rounded Rect [id:dp039237545843641586] 
\draw  [color={rgb, 255:red, 155; green, 155; blue, 155 }  ,draw opacity=1 ] (420,199.2) .. controls (420,168.71) and (444.71,144) .. (475.2,144) -- (640.8,144) .. controls (671.29,144) and (696,168.71) .. (696,199.2) -- (696,452.8) .. controls (696,483.29) and (671.29,508) .. (640.8,508) -- (475.2,508) .. controls (444.71,508) and (420,483.29) .. (420,452.8) -- cycle ;

% Text Node
\draw (334.5,71) node [scale=1.2]  {$C( E(\mathbf{x}) ,\theta _{C})$};
% Text Node
\draw (333,241.5) node [scale=1.2]  {$D_{1}( E(\mathbf{x}) ,\theta _{1})$};
% Text Node
\draw (335,348) node [scale=1.2]  {$D_{2}( E(\mathbf{x}) ,\theta _{2})$};
% Text Node
\draw (336,454) node [scale=1.2]  {$D_{3}( E(\mathbf{x}) ,\theta _{3})$};
% Text Node
\draw (187.5,180) node [scale=1.2]  {$E(\mathbf{x} ,\phi )$};
% Text Node
\draw (493,200) node   {$y_{1} =1$};
% Text Node
\draw (599,201) node   {$y_{1} =0$};
% Text Node
\draw (495,307) node   {$y_{2} =1$};
% Text Node
\draw (497,415) node   {$y_{3} =1$};
% Text Node
\draw (593,306) node   {$y_{2} =0$};
% Text Node
\draw (593,414) node   {$y_{3} =0$};
% Text Node
\draw (61,89) node  [align=left] {Source 1};
% Text Node
\draw (59,154) node  [align=left] {Source 2};
% Text Node
\draw (61,219) node  [align=left] {Source 3};
% Text Node
\draw (57,58) node  [align=left] {Batch $\mathbf{x}$};
% Text Node
\draw (557,75) node  [align=left] {{\large Task label prediction}};
% Text Node
\draw (558,166) node  [align=left] {Empirical $\displaystyle \mathcal{H}$-divergence estimation};
\end{tikzpicture}
}
\caption{Proposed approach illustration.}
\label{fig:method}
\end{figure}

\section{Extra experiments}
\subsection{Impact of source domains diversity on unseen domain accuracy}
In this experiment, we verify whether removing examples from one source domain impacts the performance on the target domain. We evaluate each target domain on models trained using all possible combinations of the remaining domains as sources. The ERM baseline is also included for reference. Results presented in Table \ref{tab:source_div} show that for all unseen domains, decreasing the number of source domains from 3 (see Table \ref{res:tab_vlcs}) to 2 hurt the classification performance for almost all combinations of source domains. We notice that in some cases, excluding a particular source from the training severely decreases the target loss. As an example, for the Caltech-101, excluding from training examples from the VOC dataset decreased the accuracy in more than $10\%$ for the proposed approach, as well as for ERM.    

\begin{table}[h]
\centering
\caption{Impact of decreasing the number of source domains on VLCS. Rows represent the two source domains used.}
\resizebox{0.6\columnwidth}{!}{
\begin{tabular}{cc|cccccc}
\hline
\multicolumn{1}{l}{} & \multicolumn{1}{l}{} & \multicolumn{6}{|c}{Source}                    \\ \hline
Target               & Method               & VC    & VL    & VS    & LC    & LS    & CS    \\ \hline
\multirow{2}{*}{V}   & ERM                   & -     & -     & -     & 66.14 & 72.16 & 69.89 \\
                     & Ours             & -     & -     & -     & 62.39 & 69.89 & 67.23 \\
\multirow{2}{*}{L}   & ERM                   & 58.32 & -     & 62.11 & -     & -     & 59.85 \\
                     & Ours             & 65.37 & -     & 65.87 & -     & -     & 64.37 \\
\multirow{2}{*}{C}   & ERM                   & -     & 98.82 & 98.58 & -     & 84.67 & -     \\
                     & Ours             & -     & 95.75 & 96.70 & -     & 81.84 & -     \\
\multirow{2}{*}{S}   & ERM                   & 69.04 & 66.29 & -     & 59.80 & -     & -     \\
                     & Ours             & 69.54 & 68.43 & -     & 57.06 & -     & -     \\ \hline
\end{tabular}}
\label{tab:source_div}
    \label{tab:my_label}
\end{table}

\subsection{Effect of random projection size}
\begin{figure}[h!]
    \centering
    \includegraphics[width=0.5\columnwidth]{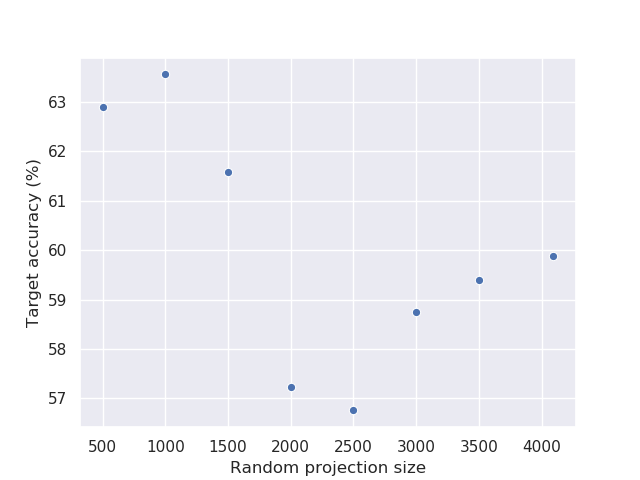}
    \caption{Accuracy obtained on the PACS benchmark using Sketch as target domain.}
    \label{fig:res_rpsize}
\end{figure}

We further investigate the effectiveness on providing a more stable training of the random projection layer in the input of each discriminator. For that, we run experiments with 7 different projection sizes, as well as directly using the output of the feature extractor model. Besides the random projection size, we use the same hyperparameters values (the same used in the previous experiment) and initialization for all models. We report in Figure \ref{fig:res_rpsize} the best target accuracy achieved with all random projection sizes on the PACS benchmark considering the Sketch dataset as unseen domain. Overall, we observed that the random projection layer has indeed an impact on the generalization of the learned representation and that the best result was achieved with a size equal to 1000. Moreover, we notice that, in this case, having a smaller (500) random projection layer is less hurtful for the performance than using a larger one. We also found that removing the random projection layer did not allow the training to converge with this experimental setting.

\section{Domain generalization benchmarks}
The VLCS benchmark is composed by 4 datasets with 5 common classes, namely, bird, car, chair, dog, and person. The number of data points per dataset is detailed as follows. We split each dataset in 80$\%$/20$\%$ train/test partitions. 
\begin{itemize}
    \item Pascal VOC2007: 3376;
    \item LabelMe: 2656;
    \item Caltech-101: 1415;
    \item SUN09: 3282.
\end{itemize}
The PACS benchmark is composed by 4 datasets with 7 common classes, namely, dog, elephant, giraffe, guitar, horse, house, and person. The number of data points per dataset is detailed as follows. We use the original train/validation partitions provided by the benchmark authors.
\begin{itemize}
    \item Photos: 1670;
    \item Art painting: 2048;
    \item Cartoon: 2344;
    \item Sketch: 3929.
\end{itemize}

\section{Implementation details}
\subsection{VLCS and PACS benchmarks}
In order to obtain a consistent comparison with the aforementioned baseline models, we follow previous work and employ the weights of a pre-trained AlexNet \cite{krizhevsky2012imagenet} and ResNet-18 \cite{he2016deep}  as the initialization for the feature extractor model on the experiments. \iffalse  with both benchmarks \fi The last layer is discarded and the representation of size 4096 for AlexNet and 512 for ResNet-18 is used as input for the task classifier and the domain discriminators. The domain discriminator architecture with AlexNet, consists of a four-layer fully-connected neural network of size $4096 \rightarrow \text{random projection size} \rightarrow 1024 \rightarrow 1$ and five-layer fully connected network of size $512 \rightarrow \text{random projection size} \rightarrow 512 \rightarrow 256 \rightarrow 1 $ for ResNet-18. The random projection layer is implemented as a linear layer with weights normalized to have unitary L2-norm. The task classifier is a one-layer fully-connected network of size $4096 \rightarrow$ number of classes in the case of AlexNet and $512 \rightarrow$ number of classes in the case of ResNet.
Following previous work on domain generalization \cite{li2017deeper, li2019episodic}, we use models pre-trained on the ILSVRC dataset \cite{ILSVRC15} as initialization. For fair comparison, all models we implemented were given a budget of 200 epochs. We use label smoothing \cite{szegedy2016rethinking} on the task classifier in order to prevent overfitting. Models were trained using SGD with Polyak's acceleration. One epoch corresponds to the length of the largest source domain training sample. The learning rate was ``warmed-up'' for a number of training iterations equal to $nw$. Hyperparameter tuning was performed through random search over a pre-defined grid so as to find the best values for the learning rate (lr), momentum, weight decay, label smoothing parameter $ls$, $nw$, random projection size\footnote{The option of not having the random projection layer is included in the grid search.},
learning rate reduction factor, and weighting ($\alpha$). Each model was run with three different initializations (random seeds 1, 10, and 100 selected \textit{a priori}) and the average best accuracy on the test partition of the target domain is reported. Details of the hyperparameters grid used in the search are provided in the Supplementary material. For our ERM we used the same hyperparameters as in \cite{carlucci2019domain}, while for IRM we employed the same hyperparameter values reported in the authors implementation of the colored MNIST experiments.

The grids used on the hyperparameter search for each hyperparameter are presented in the following. A budget of 200 runs was considered and for each combination of hyperparameters each model was trained for 200 and 30 epochs in the case of AlexNet and ResNet-18, respectively. The best hyperparamters values for AlexNet on PACS and VLCS benchmarks are respectively denoted by  $^*$, $^{\dagger}$. For the ResNet-18 experiments on PACS we indicate the hyperparameters by $^{+}$. Moreover, in the case of ResNet-18, we aggregated the discriminators losses by computing the corresponding hypervolume as in \cite{albuquerque2019multi}, with a nadir slack equal to 2.5. All experiments were run considering a minibatch size of 64 (training each iteration took into account 64 examples from each source domain) on single GPU hardware (either an NVIDIA V100 or NVIDIA GeForce GTX 1080Ti).

\begin{itemize}
    \item Learning rate for the task classifier and feature extractor: $\{0.01^{*, +}, 0.001^{\dagger}, 0.0005\}$;
    \item Learning for the domain classifiers: $\{0.0005^{*}, 0.001, 0.005^{\dagger, +}\}$;
    \item Weight decay: $\{0.0005^{*}, 0.001, 0.005^{\dagger +}\}$;
    \item Momentum: $\{0.5, 0.9^{*, \dagger, +}\}$
    \item Label smoothing: $\{0.0^{+}, 0.1, 0.2^{*, \dagger}\}$;
    \item Losses weighting ($\alpha$): $\{0.35, 0.8^{*, \dagger, +}\}$;
    \item Random projection size: $\{1000^{*}, 3000, 3500^{\dagger}, \text{None}^{+} \}$;
    \item Task classifier and feature extractor learning rate warm-up iterations: $\{1, 300^{*, \dagger}, 500^{+} \}$;
    \item Warming-up threshold: $\{0.00001^{*}, 0.0001^{\dagger, +}, 0.001\}$;
    \item Learning rate schedule patience: $\{25^ {+}, 60^{\dagger}, 80^{*}\}$;
    \item Learning rate schedule decay factor: $\{0.1 ^ {+}, 0.3^{\dagger}, 0.5^{*}\}$.
\end{itemize}

\subsection{Affective state prediction}
We use SyncNet \cite{li2017targeting} as the encoder for the experiments with the SEED dataset. We follow previous work and apply a simple pre-processing that consists of clipping artifacts with amplitude 5 times higher than the mean of the channel signal and windowing data with chunks of 60 seconds. Each window was normalized to have zero mean and unit variance. For the encoder network, we adopt an one layer parameterized convolutional filter with 2 filters (designed to extract synchrony coherence which interpretable features based on the previous neuroscience literature \cite{li2017targeting}). %The output of the encoder 602 is used as input for the task classifier and the domain discriminators. 
We train all models for 100 epochs using SGD with Polyak’s acceleration. The learning rate was ``warmed-up'' for a number of training iterations equal to 500.

The output of the encoder with size 602 is used as input for the task classifier and the domain discriminators. The domain discriminator architecture consists of a four-layer fully-connected neural network of size $602 \rightarrow \text{random projection size} \rightarrow 256  \rightarrow 128 \rightarrow 2$. The random projection layer is implemented as a linear layer with weights normalized to have unitary L2-norm. The task classifier is a two-layer fully-connected network of size $602 \rightarrow 100 \rightarrow$ number of classes.

The summary of parameters is presented in the following.
\begin{itemize}
    \item Window size: 60 seconds 
    \item Number of filters: 2 
    \item Filters length: 40
    \item Pooling  size: 40 
    \item Input drop out rate: 0.2 
    \item Initial learning rate task classifier: 9.963e-04
    \item Initial learning rate discriminator: 9.963e-05
    \item Random projection size: 602
\end{itemize}

\subsection{Proxy $\mathcal{A}$-distance estimation}
We implement the domain discriminators using tree ensemble classifiers with 100 estimators. We thus report the average classification accuracy using 5-fold cross-validation independently run for each domain pair.  Each domain is represented by a random sample of size 500.

\end{document}